\documentclass[letterpaper]{article} 
\usepackage{aaai2026}  
\usepackage{times}  
\usepackage{helvet}  
\usepackage{courier}  
\usepackage[hyphens]{url}  
\usepackage{graphicx} 
\usepackage{natbib}  
\usepackage{caption} 
\usepackage{algorithm}
\usepackage{algorithmic}
\usepackage{amsmath}
\usepackage{booktabs}
\usepackage{amssymb}

\usepackage{newfloat}
\usepackage{listings}
\DeclareCaptionStyle{ruled}{labelfont=normalfont,labelsep=colon,strut=off} 
\floatstyle{ruled}
\newfloat{listing}{tb}{lst}{}
\floatname{listing}{Listing}
\title{GloTok: Global Perspective Tokenizer for Image Reconstruction and Generation}
\author{
    Xuan Zhao\textsuperscript{\rm 1}\equalcontrib \thanks{This work was done by Xuan Zhao during an internship at Tencent Youtu Lab.},
    Zhongyu Zhang\textsuperscript{\rm 2}\equalcontrib,
    Yuge Huang\textsuperscript{\rm 2},
    Yuxi Mi\textsuperscript{\rm 1},
    Guodong Mu\textsuperscript{\rm 2},
    Shouhong Ding\textsuperscript{\rm 2}\thanks{Corresponding authors.},
    Jun Wang\textsuperscript{\rm 3},
    Rizen Guo\textsuperscript{\rm 3},
    Shuigeng Zhou\textsuperscript{\rm 1}\footnotemark[3]
}

\affiliations{
    \textsuperscript{\rm 1}College of Computer Science and Artificial Intelligence, Fudan University, Shanghai, China\\
    \textsuperscript{\rm 2}Tencent Youtu Lab, Shanghai, China\\
    \textsuperscript{\rm 3}Tencent WeChat Pay Lab33, Shanghai, China\\
    xzhao23@m.fudan.edu.com,
    \{yxmi20, sgzhou\}@fudan.edu.com\\
    \{wilxyzhang, yugehuang, gordonmu, ericshding\}@tencent.com \\
    \{earljwang, rizenguo\}@tencent.com\\
}

\usepackage{bibentry}

\begin{document}

\maketitle

\begin{abstract}
Existing state-of-the-art image tokenization methods leverage diverse semantic features from pre-trained vision models for additional supervision, to expand the distribution of latent representations and thereby improve the quality of image reconstruction and generation.
These methods employ a locally supervised approach for semantic supervision, which limits the uniformity of semantic distribution. However, VA-VAE proves that a more uniform feature distribution yields better generation performance.
In this work, we introduce a Global Perspective Tokenizer (GloTok), which utilizes global relational information to model a more uniform semantic distribution of tokenized features.
Specifically, a codebook-wise histogram relation learning method is proposed to transfer the semantics, which are modeled by pre-trained models on the entire dataset, to the semantic codebook.
Then, we design a residual learning module that recovers the fine-grained details to minimize the reconstruction error caused by quantization.
Through the above design, GloTok delivers more uniformly distributed semantic latent representations, which facilitates the training of autoregressive (AR) models for generating high-quality images without requiring direct access to pre-trained models during the training process.
Experiments on the standard ImageNet-1k benchmark clearly show that our proposed method achieves state-of-the-art reconstruction performance and generation quality. 
\end{abstract}


\section{Introduction}

Recent years have witnessed remarkable advancements in image generation, driven by autoregressive (AR) models~\cite{DBLP:journals/corr/VaswaniSPUJGKP17, radford2018improving, touvron2023llama} and diffusion models~\cite{ dhariwal2021diffusion, ho2022classifier, rombach2022high}.
To promote high-quality generation, many contemporary methods adopt a two-stage paradigm:
(1) a tokenizer compresses input images into tokens, followed by (2) a generator modeling the distribution of tokens.
The compressed tokens determine both structural composition and detailed characteristics of generated images.
Therefore, image tokenizers play a crucial role in improving the performance of image generation.

\begin{figure}[t]
  \centering
  \includegraphics[width=0.9\linewidth]{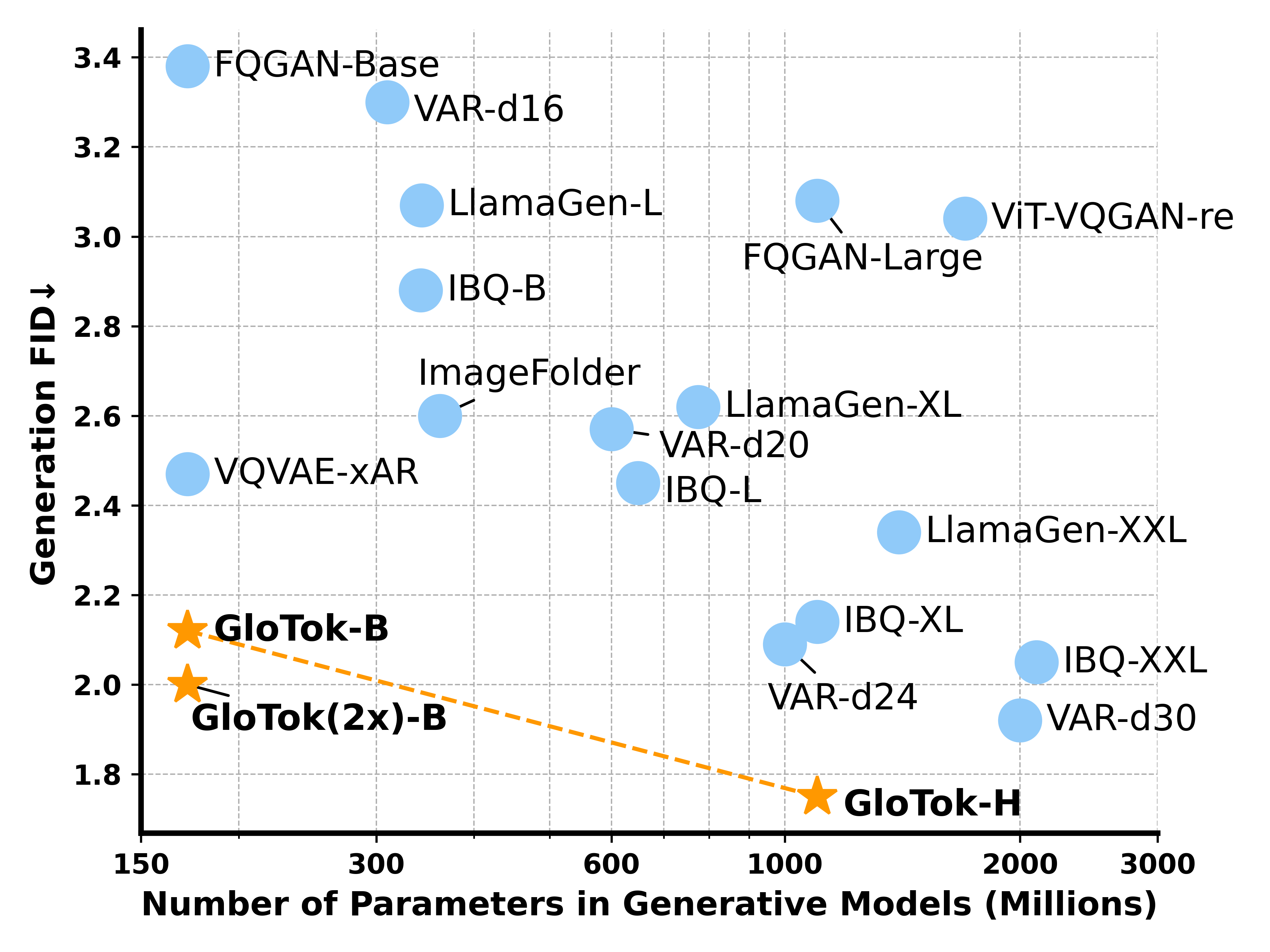} 
  \caption{Comparison of generative performance among different methods and GloTok, where lower values on the Y-axis correspond to better performance.}
  \label{img:gfid}
\end{figure}

Recent SOTA tokenizers~\cite{zhu2024stabilize,qu2024tokenflow,li2024imagefolder,bai2024factorized,ma2025unitok} for image generation with AR models leverage pre-trained models such as DINO~\cite{DBLP:journals/corr/abs-2104-14294,oquab2023dinov2, darcet2023vitneedreg} and CLIP~\cite{radford2021learning} as additional supervision.
Concretely, the pre-trained model converts the input images into semantic features, and these features are then served to guide the tokenizer to generate more effective latent representations for image generation.

While these approaches have achieved remarkable performance in enhancing the information capacity of latent representations, their dependence on single-image semantic supervision poses a significant challenge to semantic codebook fitting.
Specifically, this local approach is confined to performing contrastive learning between latent-space features and pre-trained features within individual images, thereby constraining its capacity to model feature discrepancies across the entire dataset. Consequently, the learned latent-space distribution exhibits suboptimal breadth and uniformity.
VA-VAE~\cite{yao2025reconstruction} demonstrates that more uniform latent representations can significantly improve generative performance.
Based on the conclusions of this prior work, suboptimal uniform feature distributions can tend to constrain generative performance.

In this paper, we introduce GloTok, a novel tokenizer that outperforms previous work in Fig.~\ref{img:gfid} and is designed to achieve a more uniform latent distribution, leveraging global relations across the entire dataset.
In practice, GloTok adopts a dual codebook architecture to learn the semantics and details of the input images. A codebook-wise histogram relation learning method is proposed to transfer the semantics to the semantic codebook by feature relationships modeled by pre-trained models across the entire dataset (called global relations).
Taking ImageNet as an example, existing methods require batch-based semantic learning for 1.28 million images, resulting in unstable optimization and makes it difficult to achieve global optimality.
Differently, our approach consistently leverages global relations as the semantic guidance, such as a multi-bin histogram to summarize the global relations.
This design not only facilitates improved uniformity of semantic latent distribution, but also eliminates the need to directly access the pre-trained model during training, thereby reducing both temporal and spatial overhead.

In addition, to mitigate accuracy degradation caused by discretization, we employ a residual learning module in GloTok.
This module takes quantized features as input, predicts the residual between the original continuous features and their quantized discrete counterparts, and fuses this residual with the quantized features to produce the final quantized feature representations.
Residual learning enables the tokenizer to partially restore fine-grained details, thereby improving the quality of image reconstruction.

Through a novel histogram relation learning method and residual learning modules, the performance of the AR model trained with our proposed GloTok achieves significant improvements in generative tasks.
Our experiments demonstrate that the proposed model achieves a state-of-the-art (SOTA) reconstruction FID of 0.83 at 256$\times$256 resolution on the standard ImageNet benchmark.
And we get a competitive generation FID=1.75 on the ImageNet 256$\times$256 image generation using an autoregressive generator.

Our contribution can be summarized as follows.
\begin{itemize}
    \item We propose GloTok, a novel tokenizer that constrains the global relational distribution within the latent space to transfer the semantics from pre-trained models, achieving a significant improvement in the performance of image reconstruction and generation.
    \item A codebook-wise histogram relation learning method is introduced to transfer the semantics to the semantic codebook, which models a more uniform latent distribution and eliminates the need to directly access the pre-trained models during training.
    \item To mitigate accuracy degradation induced by discretization, a residual learning module is designed to predict and fuse token residuals into the final representations, preserving fine-grained details for superior image quality.
\end{itemize}

\section{Related Work}
\subsection{Image Tokenization}
Image tokenization is the fundamental process in image reconstruction and generation. An image tokenizer converts the input image into continuous or discrete tokens.

As a pioneering work on transformer-based image generation, VQ-GAN~\cite{esser2021taming}  proposes the encoder-quantizer-decoder architecture, which employs a vector quantization~(VQ) technique to map the continuous latent space features of the input images into discrete tokens.
However, the performance of VQGAN is constrained by both a small codebook size and a low token utilization rate.
ViT-VQGAN~\cite{yu2021vector} and Efficient-VQGAN~\cite{cao2023efficient} leverage the Vision Transformer (ViT) architecture~\cite{dosovitskiy2020image}, which mitigates reliance on convolutional inductive biases due to its global attention mechanisms.
RQ-VAE~\cite{lee2022autoregressive} adopts a residual quantization (RQ) to precisely approximate the feature map.
VQGAN-LC~\cite{zhu2024scaling} and SimVQ~\cite{zhu2024addressing} employ linear layers to expand the codebook capacity while preserving high codebook utilization efficiency.
Furthermore, MAGViT-v2~\cite{yu2023language} proposes a lookup-free quantization method leading to a large vocabulary.
Unlike typical 2D tokenizers, TiTok~\cite{yu2024image} adopts a 1D tokenization framework that enables a more flexible token count design while also capturing semantically rich image information.

Recent advancements have focused on injecting semantic features from pre-trained vision models such as DINO~\cite{DBLP:journals/corr/abs-2104-14294,oquab2023dinov2, darcet2023vitneedreg} or CLIP~\cite{radford2021learning} into the codebook to enhance the generation performance of VQGAN.
DiGiT~\cite{zhu2024stabilize} integrates the discrete tokens derived from clustered DINO~\cite{oquab2023dinov2} features into the latent space of VQGAN to enrich the semantic information of the latent space of the generative model.
UniTok~\cite{ma2025unitok} adopts a pre-trained CLIP~\cite{radford2021learning} model to align the latent space of input images with the semantic latent space derived from the corresponding text.
Recent work, such as ImageFolder~\cite{li2024imagefolder} and FQGAN~\cite{bai2024factorized}, utilizes a multi-codebook architecture in which certain codebooks are designed to capture the feature distributions extracted from pre-trained vision models using a feature-wise contrastive loss. 
These methods adopt a local perspective, specifically by capturing semantics through intra-image feature comparison with pre-trained models.
In contrast, our tokenizer GloTok employs a method that globally constrains the distribution of semantic features with pre-trained relational distributions to transfer the semantics.

\subsection{AR Image Generation}
Recent years have witnessed remarkable advancements in auto-regressive (AR) image generation, where models synthesize images by capturing the distribution of discrete tokens.
Early foundational works, such as VQGAN~\cite{esser2021taming}, VIT-VQGAN~\cite{yu2021vector}, and RQ-Transformer~\cite{lee2022autoregressive}, employ transformer architecture to model discrete token sequences for image generation.
Building on this, LlamaGen~\cite{sun2024autoregressive} introduces the Llama~\cite{touvron2023llama} model to enhance image generation capabilities. Beyond architectural innovations, optimizing token training strategies has emerged as a key research direction. For instance, VAR~\cite{tian2024visual} adopts a multi-resolution token prediction framework to improve the image fidelity and detail preservation.
MAR~\cite{li2024autoregressiveimagegenerationvector} models the per-token probability distribution using a diffusion procedure, which achieves strong generation results.
FlowAR~\cite{ren2024flowar} introduces a general next-scale prediction method with a streamlined scale design.
RAR~\cite{yu2024randomized} randomly permutes the input token sequence into different factorization orders for modeling bidirectional contexts.
xAR~\cite{ren2025nexttokennextxpredictionautoregressive} enables flexible prediction units and effectively alleviates exposure bias, achieving significant generative performance.

\section{Method}
\subsection{Preliminary}
AR image generation models are trained from discrete tokens generated from an image tokenizer.
The tokenizer typically follows an encoder-quantizer-decoder architecture, where the encoder extracts features from the image, the quantizer converts these features into discrete tokens, and the decoder reconstructs the image from these tokens. This process allows the model to effectively learn and generate images based on tokenized representations.

Given an image $x \in \mathcal{R}^{H\times W\times 3} $, the encoder $E$ first extracts continuous features $Z^{h\times w\times c}$ from the input image $x$, where $h \times w $ represents the sequence length of the latent tokens and c represents the channel dimension.
For image quantization, the quantizer selects the closest token $t$ for each feature $z$ of $Z$ with a learnable codebook $ C^{n\times d} $ by calculating the Euclidean distance~\cite{esser2021taming} or the cosine similarity~\cite{shi2024taming, unitok, bai2024factorized}. The quantization loss $L_{Q}$ is calculated based on the quantized features $\hat{Z}^{h\times w \times c}$ and the original features $Z$ to optimize the latent space and the codebook:
\begin{equation}
L_{Q} = \left \| sg \left [ Z \right ] - \hat{Z} \right\| _{2}^{2} + \beta \left \| sg \left [ \hat{Z} \right ] - Z \right\| _{2}^{2}
\end{equation}
Here, sg[·] denotes the stop-gradient operation and $ \| \cdot \|_{2}^{2} $ denotes the Euclidean distance (or called L2 distance).
The quantization process optimizes latent representations within the codebook while reducing the representation gap between the continuous latent space and the discrete latent space.

After quantization, the decoder $D$ decodes $\hat{Z}$ into the reconstruction image $\hat{x}$. 
For the representative VQGAN~\cite{esser2021taming} model, the optimization of the entire network is represented by the following loss function $L_{VQGAN}$:
\begin{equation}
L_{R} = \left \| x - \hat{x} \right\| ^{2}
\end{equation}
\begin{equation}
     L_{VQGAN} = \lambda_{R}L_{R} + \lambda _{Q}L_{Q} + \lambda_{P}L_{P} + \lambda_{G}L_{G} 
\end{equation}
where $\lambda_i $ is the weight factor for each loss term, $L_R$ is the reconstruction loss, $L_{P}$ is the perceptual loss between $x$ and $\hat{x}$, and $L_{G} $ is adversarial loss encouraging to learn more realistic images $x$ with a patch-discriminator ~\cite{isola2017image}.

\begin{figure}[t]
  \centering
  \includegraphics[width=\linewidth]{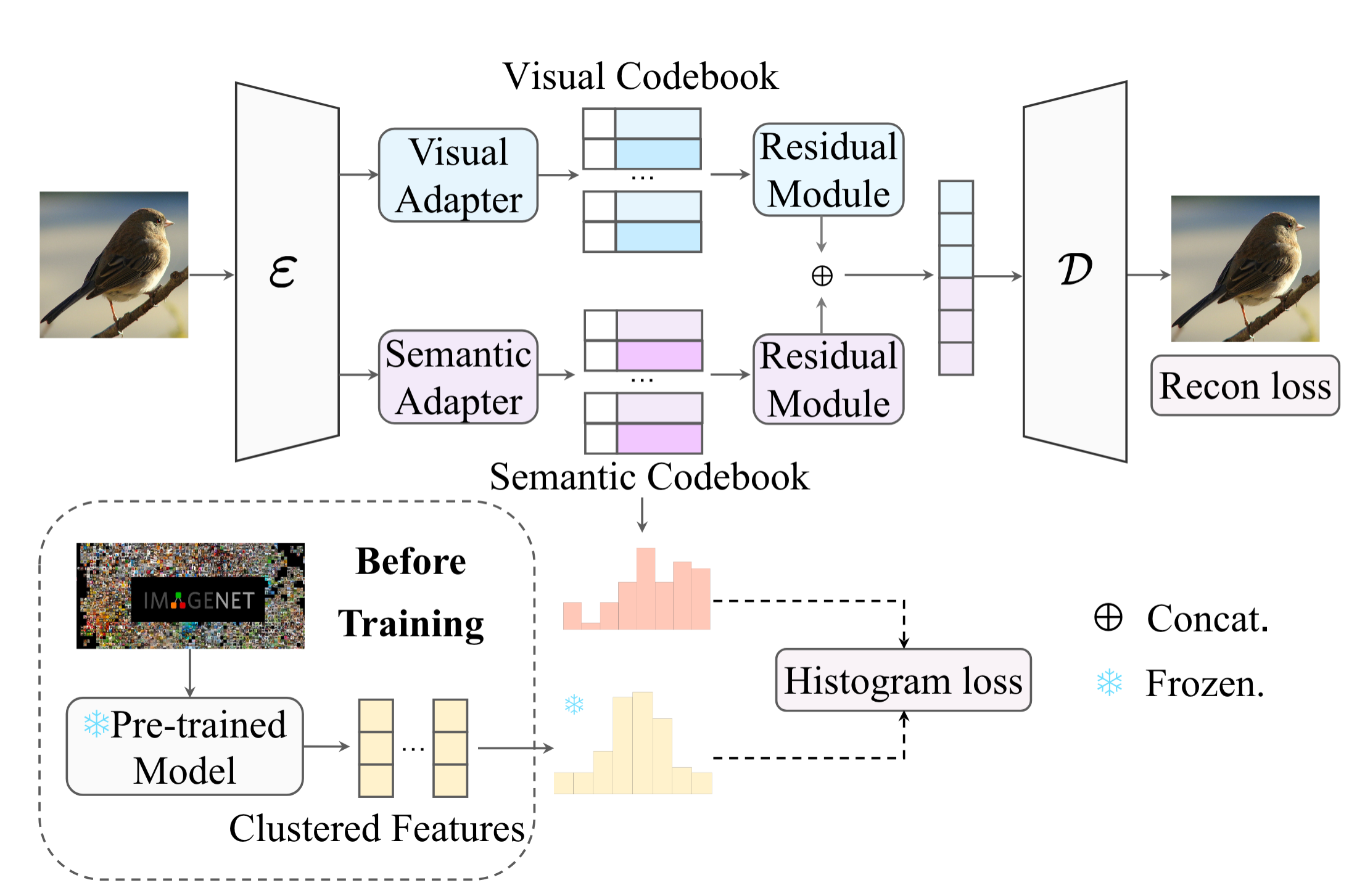}
  \caption{Illustration of our method. Top: GloTok encoder-quantizer-decoder architecture with dual codebooks and residual modules. Bottom: overview of the Histogram Relation Learning method. An image is quantized into two sets of tokens by a visual codebook and a semantic codebook. The semantic codebook learns the token relationship from features clustered from a pre-trained model with a histogram loss. GloTok adopts two residual modules to learn the residuals between continuous features and discrete features.}
  \label{framework}
\end{figure}

\subsection{Architecture}
Similarly to previous methods~\cite{bai2024factorized, li2024imagefolder}, as shown in Fig.~\ref{framework}, we leverage a dual-codebook architecture with a convolutional encoder and decoder.
we encode the input image into two sets of tokens: one consisting of the semantic tokens and the other of the visual tokens. 
We use the semantic tokens to model the distribution of semantic features, while visual tokens are employed to model the image's visual features.
Specifically, the encoder $E$ first encodes the input image $x$ into the latent features $Z^{ h \times w \times c}$. We then adopt two adapters to decompose $Z$ into semantic features $Z_{sem}$ and visual features $Z_{vis}$.
For quantization, $Z_{sem}$ and $Z_{vis}$ are quantized into $\hat{Z}_{sem}$ and $\hat{Z}_{vis}$ respectively, with learnable codebooks $C_{sem}$ and $C_{vis}$. 

To control the global feature space distribution of semantic features, we introduce a novel codebook-wise histogram relation learning method that constrains the distribution of relations between tokens in the semantic codebook, detailed in Section~\ref{sec:our-model}.
The visual codebook is specifically served to capture visual feature representations within the latent space, enabling image reconstruction by leveraging structural patterns learned from the semantic codebook.

Quantization of the continuous latent $Z$ into discrete representations inherently induces information degradation. To mitigate the limitation, we designed a residual learning module to recover lost information, thus improving the quality of image reconstruction.
The learned residual features are aggregated with quantized tokens through summation operations to generate information-complete token representations.
Subsequently, the decoder decodes the composite latent representation $\hat{Z}$ formed by channel-wise concatenation of $\hat{Z}_{sem}$ and $\hat{Z}_{vis}$ to synthesize the output image.

\subsection{GloTok}
\label{sec:our-model}
\subsubsection{Histogram relation learning}
In order to visualize the distribution of the quantitative data, the histogram is a commonly used method that reflects the probability density of the underlying data distribution.
Inspired by this concept, we propose a codebook-wise histogram relation learning method that transfers the prior global relational distribution knowledge to the semantic codebook. 

To achieve this objective, it is necessary to select an appropriate source for the prior distribution.
We adopt two approaches for its construction: (1) directly utilizing the semantic codebook of the tokenizer from a pre-trained model; (2) performing k-means clustering on the intermediate layer features of CLIP~\cite{radford2021learning} or DINOv2~\cite{oquab2023dinov2} adopted in VQGAN-LC~\cite{zhu2024scaling}.

In particular, our methodology differs from VQGAN-LC in that it employs an MLP layer to establish the mapping from clustered features to the codebook.
Although the feature sources in VQGAN-LC contain rich semantic information, the semantic constraints imposed on the mapped codebook are insufficient, and the use of an MLP layer for training a large codebook, which is up to 100,000, results in substantial computational cost.
Instead, we focus on the patterns of relational similarity between tokens.

To construct the global relational distribution, we first compute the pairwise cosine similarities between tokens within each token set $C^{n\times d}$:
\begin{equation}
  D = \{ \left. f(t_i,t_j) \right| t_i,t_j \in C,i\le j \}
\end{equation}
where f[$\cdot$] is a \textit{cosine similarity} function.
Subsequently, in a histogram-like manner, we initialize N bins slicing from -1 to 1, where each bin corresponds to a specific sub-range of quantized token distances:
\begin{equation}
  B = \left\{ \left.-1 + \frac{2k}{N-1} \right| k=0,1,2,...,N-1 \right\}
\end{equation}

To quantify the distribution of $D$ across different intervals $B$, we compute a Gaussian-smoothed histogram of pairwise cosine similarities:
\begin{equation}
    p_n = \sum_{i=1}^{\| D \|}\exp{(-\alpha(D_i-B_n)^{2})}
\end{equation}
where $\| D\|$ is the dim of D and $\alpha$ is a weight factor.

We define the relational distribution of the trainable codebook as the student distribution $P_s$ and define the relational distribution of pre-trained tokens as the teacher distribution $P_t$.
Then, we employ a Kullback-Leibler divergence to constrain the student distribution with the teacher distribution:
\begin{equation}
    L_{hist} = KL( \log{P_{s}} \| sg\left [ P_{t} \right])
\end{equation}
where sg[·] denotes the stop-gradient operation, $P_s$ denotes the normalized student distribution, and $P_t$ denotes the normalized teacher distribution.

\subsubsection{Residual learning} 
The quantization process, which transforms continuous features into discrete representations, inherently induces precision degradation.
While discrete representations simplify the latent space expression, they inevitably compromise the model's representational capacity.

We propose a residual learning method in which residual modules $ R_{sem} $ and $ R_{vis} $ are implemented as transformer blocks process quantized latent features $\hat{Z}_{sem}$ and $\hat{Z}_{vis}$, respectively, to predict lost latent details during discretization:
\begin{equation}
    Z_{res,sem} = R_{sem}(\hat{Z}_{sem})
\end{equation}
\begin{equation}
    Z_{res,vis} = R_{vis}(\hat{Z}_{vis})
\end{equation}
The optimization is derived from the differential mapping between continuous features $Z$ and the quantized features $\hat{Z}$ through residual error minimization:
\begin{equation}
\begin{split}
    L_{res} & = \left \| sg\left [ Z_{sem} \right ] - \hat{Z}_{sem} - Z_{res,sem}  \right \| \\
            & + \left \| sg\left [ Z_{vis} \right ] - \hat{Z}_{vis} - Z_{res,vis}  \right \|
\end{split}
\end{equation}
Discrete features$\hat{Z}_{sem}$ and $\hat{Z}_{vis}$ are combined with their respective learned residuals through an element-wise addition operation to form the final discrete representation:
\begin{equation}
    \hat{Z}_{vis} = \hat{Z}_{vis} + Z_{res,vis}
\end{equation}
\begin{equation}
    \hat{Z}_{sem} = \hat{Z}_{sem} + Z_{res,sem}
\end{equation}

\subsubsection{Training loss}
Our model is trained with a combined loss:
\begin{equation}
\begin{aligned}
    L =  L_{VQGAN} + \lambda _{res}L_{res} + \lambda _{hist}L_{hist}
\end{aligned}
\end{equation}
where $\lambda _{res}$ and $\lambda _{hist}$ are weight factor that respectively control $L_{res}$ and $L_{hist}$, thus balancing various loss terms.

\subsection{Training with Auto-Regressive Model}
Through quantization, the input image is discretized into two complementary token sets in our tokenizer. Previous auto-regressive models~\cite{esser2021taming, touvron2023llama, tian2024visual} are trained with the indices of the tokens quantized by codebook. These auto-regressive models are designed to model inter-token dependencies via sequential pattern learning.

However, although previous multi-branch models~\cite{bai2024factorized, li2024imagefolder}  exhibited superior reconstruction performance compared to other single-branch methods~\cite{sun2024autoregressive, tian2024visual}, their generation performance remained relatively suboptimal.
When the tokenizer employs a multi-branch architecture, it must generate multiple indices to represent images.
Due to the exposure bias~\cite{arora2023exposurebiasmattersimitation} inherent in AR training, producing multiplicative indices exacerbates index prediction errors during inference, thus degrading generation performance. In our work, we aim to circumvent this limitation. 

We concatenate the semantic features and visual features obtained after vector quantization along the channel dimension, and employ an xAR~\cite{ren2025nexttokennextxpredictionautoregressive} model to directly learn this combined feature space.
This approach enhances the generation performance of the multi-branch tokenizer.

\begin{table*}[ht]
  \small
  \centering
  \begin{tabular}{c|cc|ccc}
    \toprule
    \textbf{Codebook Type} &\textbf{ Method} &\textbf{Codebook Size} &\textbf{rFID $\downarrow$}  & \textbf{PSNR $\uparrow$} \\
    \midrule
    Single & VQGAN~\cite{esser2021taming} & 16,384 & 5.96 & 23.3  \\
    Single & RQ-VAE~\cite{lee2022autoregressive} & 16,384  & 3.20 & - \\
    Single & VQGAN-LC~\cite{zhu2024scaling} & 16,384  & 3.01 & 23.2  \\
    Single & VQGAN-LC~\cite{zhu2024scaling} & 100,000  & 2.62 & 23.8  \\
    Single & LlamaGen~\cite{sun2024autoregressive} & 16,384  & 2.19 & 20.7  \\ 
    Single & MaskBit~\cite{weber2024maskbit} & 16,384  & 1.37 & 21.5 \\
    Signle & TiTok-S-128~\cite{yu2024image} & 4,096  & 1.71 & 17.7 \\
    Single & Open-MAGVIT2~\cite{luo2024open} & 16,384  & 1.58 & -\\
    Single & Open-MAGVIT2~\cite{luo2024open} & 262,144  & 1.17 & 22.6  \\
    Single & IBQ~\cite{shi2024taming} & 16,384  & 1.37 & 21.0 \\
    Single & IBQ~\cite{shi2024taming} & 262,144  & 1.00 & -  \\
    \midrule
    Muti & TokenFlow~\cite{qu2024tokenflow} & $32,768 \times 2$  & 1.37 & 21.4 \\
    Muti & FQGAN-Dual~\cite{bai2024factorized} & $16,384 \times 2$  & 0.94 & 22.0  \\
    \midrule
    Muti & GloTok & $16,384$ &  0.92 & 22.4 \\
    Muti & GloTok(2x) & $16,384\times2$ &  \textbf{0.83} & 22.7 \\
    \bottomrule
  \end{tabular}
   \caption{Reconstruction comparison with other tokenizers evaluated on ImageNet 256×256 50k validation dataset.}
   \label{tab:recon}
\end{table*}

\section{Experiments}
\subsection{Experiments Setting}
\subsubsection{Implementation details}
We trained and evaluated GloTok on the $256 \times 256$ ImageNet~\cite{deng2009imagenet} benchmark. 
The encoder and decoder settings are consistent with VQGAN~\cite{esser2021taming}.
To facilitate alignment in methods employing single-branch codebook and multi-branch codebook settings with the same codebook size, respectively, we trained two tokenizers with distinct codebook size configurations: (1) the semantic codebook is configured with a size of 4,096, and the visual codebook with a size of 12,288, resulting in a total size of 16,384; (2) both the semantic and visual codebooks have a size of 16,384, which we refer as GloTok(2x).
The dimension of the codebooks is 8.
To transfer inter-token relational knowledge from pre-trained models, we train the tokenizer with diverse teacher distributions, such as K-means-clustered DINOv2 tokens and a pre-trained semantic codebook. We configure the number of bins (B) of the histogram learning method to 40.
We set $\lambda _{hist} =0.01$, $\lambda _{res}=0.5$.
Within the residual learning module, we incorporate a single-layer transformer block to predict residual components.
The tokenizer is trained with the following settings: a learning rate of 2e-4 without any decay mechanism, an Adam Optimizer with $\beta_1$ = 0.9,  $\beta_2$ = 0.95,
a global batch size of 240 with 150 epochs.
For the generator, we train FAR-B, FAR-L with 200 epochs with our tokenizer, and we train xAR-B, xAR-H ~\cite{ren2025nexttokennextxpredictionautoregressive}  with 800 epochs following xAR settings.

\subsubsection{Metrics}
We employ the Frechet Inception Distance (FID)~\cite{DBLP:journals/corr/HeuselRUNKH17} and the Peak Signal-to-Noise Ratio (PSNR) to evaluate reconstruction quality. We use the ImageNet validation set, consisting of 50k samples, to compute the reconstruction FID (rFID) and PSNR.
To evaluate the quality of generated samples, we employ two widely recognized metrics: generation FID (gFID) and Inception Score (IS)~\cite{salimans2016improved}. The gFID measures the similarity between the distribution of generated images and that of real-world data, providing a quantitative assessment of the realism and diversity of the outputs. On the other hand, the IS evaluates both the quality and diversity of the generated samples by leveraging an Inception model pretrained on the ImageNet dataset. 

\begin{table*}[ht]
  
  \centering
  \small
  \begin{tabular}{c|ccc|cc}
    \toprule
    \textbf{Type} & \textbf{Method} & \textbf{Params} &  \textbf{Tokens} &\textbf{gFID$\downarrow$} & \textbf{IS$\uparrow$}\\
    \midrule
    AR & VQGAN~\cite{esser2021taming}   & 227M & 256 & 18.65 & 80.4\\
    AR & VQGAN-LC~\cite{zhu2024scaling}  & 404M & 256 & 15.4 & -\\
    AR & LlamaGen-B~\cite{sun2024autoregressive}   & 111M & 256 & 5.46 & 193.61\\
    AR & LlamaGen-XXL~\cite{sun2024autoregressive}   & 1.4B & 256 & 3.09 & 253.60\\
    AR & LlamaGen-3B~\cite{sun2024autoregressive}   & 3.1B & 256 & 3.06 & 279.71\\
    AR & IBQ-B~\cite{shi2024taming}  & 342M & 256&2.88 & 254.73\\
    AR & IBQ-XXL~\cite{shi2024taming}  & 2.1B & 256&2.05 & 286.73\\
    \midrule
    VAR & ImageFolder~\cite{li2024imagefolder}  & 362M & 256 & 2.60 & 295.0\\
    VAR & VAR-d16~\cite{tian2024visual} & 310M & 256 & 3.30 & 274.4\\
    VAR & VAR-d24~\cite{tian2024visual} & 1.0B & 256 & 2.09 & 312.9\\
    VAR & VAR-d30~\cite{tian2024visual} & 2.0B & 256 & 1.92 & 323.1\\
    \midrule
    FAR & FQGAN~\cite{bai2024factorized}   & 415M & 512 & 3.38 & 248.26\\
    FAR & FQGAN~\cite{bai2024factorized}  & 898M &512& 3.08 & 272.52\\
    FAR & GloTok  & 415M &512& 3.13 & 280.66\\
    FAR & GloTok(2x)  & 415M &512& 2.95 & 261.38 \\
    FAR & GloTok  & 898M &512& 2.98 & 285.34\\
    \midrule
    xAR & VQ-VAE\dag  & 172M &256& 2.47 & 277.63\\
    xAR & GloTok  & 172M &256& 2.12 & 272.16\\
    xAR & GloTok(2x)  & 172M &256& 2.00 & 269.25\\
    xAR & GloTok  & 1.1B &256& \textbf{1.75} & \textbf{327.50}\\
    \bottomrule
  \end{tabular}
  \caption{Class-conditional generation comparison on ImageNet-1k 256×256 resolution. The evaluation protocol and implementation are the same as ADM. \dag denotes that the xAR is trained with a VQ-VAE tokenizer from LlamaGen~\cite{touvron2023llama}.}
  \label{tab:gen}
\end{table*}
\subsection{Main Results}

\subsubsection{Image reconstruction}
As shown in Tab.~\ref{tab:recon}, our tokenizer was compared with other state-of-the-art models in terms of reconstruction performance.
Reconstruction performance of our tokenizer was assessed at 256$\times$256 resolution, employing 16x downsampling, on the ImageNet-1K validation set. 

Our tokenizer achieves state-of-the-art (SOTA) reconstruction performance on latent resolution 16$\times$16, attaining a reconstruction FID of 0.83 at 16384$\times$2 codebook size. Notably, the multi-codebook architecture demonstrates substantially superior reconstruction FID compared to single-codebook architectures, validating the effectiveness of learning hierarchical latent space representations across different aspects. Furthermore, our model at a total 16384 codebook size, attaining an rFID=0.92, not only outperforms tokenizers with equivalent codebook size, but also achieves superior performance relative to those with larger codebooks.
\begin{figure}[h]
  \centering
  \includegraphics[width=0.85\linewidth]{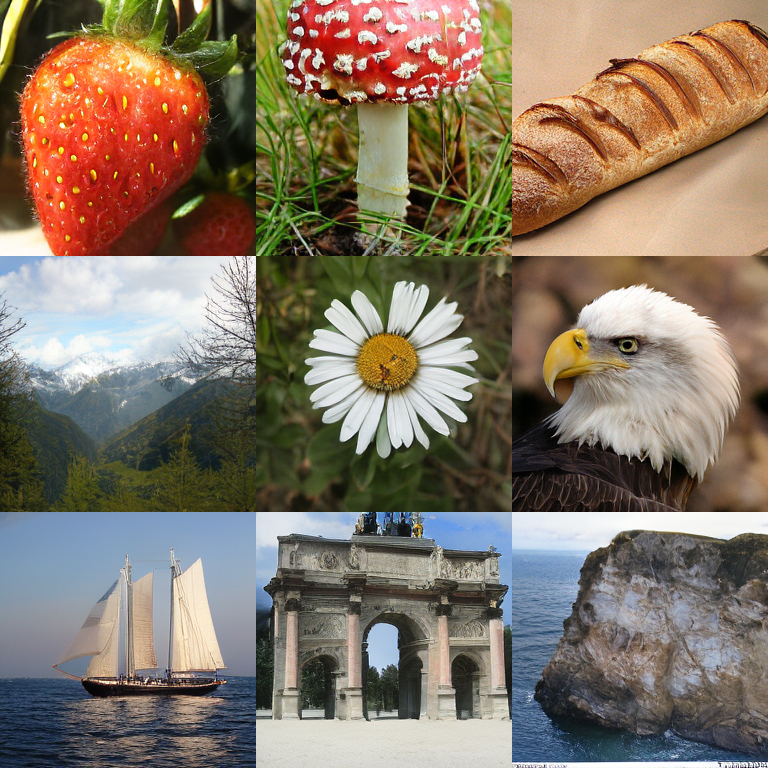}
  \caption{ImageNet-1k 256$\times$256 generated samples of GloTok trained with xAR.}
  \label{generated}
\end{figure}

\subsubsection{Image generation}
We present a comprehensive summary of the experimental results on the ImageNet-1K image generation benchmark at a resolution of 256$\times$256 pixels in Tab.~\ref{tab:gen}.
Given that the xAR method directly learns the quantized latent representations rather than relying on indices, we conducted an evaluation to assess the effectiveness of the quantized features of GloTok.
We trained xAR using GloTok configured with a total codebook size of 16384, achieving a competitive generative performance and attaining a gFID score of 1.75 and an IS of 327.50.
It should be noted that xAR gives the generation results in the VAE~\cite{kingma2022autoencodingvariationalbayes} feature space without quantization. In order to make a fair comparison with the quantization feature space, we trained xAR with a VQ-VAE-based tokenizer and achieved a gFID score of 2.47.
In comparison, when we trained xAR using GloTok with a total codebook size of 16384, the resulting gFID score dropped to 2.12, a 0.35 reduction relative to the VQ-VAE-based baseline.
Fig.~\ref{generated} shows the visualization results of our 16384 codebook size tokenizer’s 256×256 generated images training with xAR, showcasing its high fidelity.
In order to validate the impact of histogram relation learning on the generative performance, we trained the FAR-B model using a tokenizer with the same codebook size (16384$\times$2) as that of the FQGAN-dual model, which employs a local contrastive learning approach. 
Compared to FQGAN, our method achieved a gFID reduction of 0.43 and an IS improvement of 13, demonstrating that our approach yields superior generative performance relative to local contrastive methods.
These findings further validate the effectiveness of the histogram relation learning method in enhancing generative performance.

\subsection{Histogram Relation Learning Makes More Uniform}
We select quantized semantic latent representations for evaluation, like VA-VAE~\cite{yao2025reconstruction}.
We randomly select images from the Imagenet-1k val dataset. We then get the semantic latent representations $\hat{Z}_{sem}$ from FQGAN and GloTok with the same 16384$\times$2 codebook size. For VQ-VAE, we select a tokenizer from LLamaGen~\cite{touvron2023llama} with a 16384 codebook size as a baseline. The dimension of tokens from the above codebooks is all 8.
We computed the standard deviation and Gini coefficients to evaluate the uniformity. Tab.~\ref{tab:uniform} shows that the semantic latent distribution of GloTok is more uniform than FQGAN using a pre-trained model for local contrastive learning. 

As shown in Tab.~\ref{tab:gen}, the FAR trained with GloTok utilizing a codebook size of 16384$\times$2 attained a gFID score of 2.95, which surpasses the gFID score of 3.38 achieved by the FAR-B trained with FQGAN. This finding demonstrates that a more uniform semantic space distribution is more beneficial for improving generative performance.

\begin{table}[ht]
  \centering
  \small
  \begin{tabular}{c|cccc}
    \toprule
    \textbf{Tokenizer} & 
    \textbf{\shortstack[c]{density \\ cv$\downarrow$}} & 
    \textbf{\shortstack[c]{normalized \\ entropy$\uparrow$}} & \textbf{\shortstack[c]{gini \\ coefficient$\downarrow$}} \\
    \midrule
    VQ-VAE\dag & 0.221 & 0.9971 & 0.1191 \\
    FQGAN & 0.225  & 0.9970 & 0.1202\\
    GloTok & \textbf{0.217} & \textbf{0.9972} & \textbf{0.1168} \\
    \bottomrule
  \end{tabular}
  \caption{Comparison of uniformity metrics of VQ-VAE, FQGAN, GloTok. 
  \dag denotes that the VQ-VAE is obtained from LlamaGen~\cite{touvron2023llama}.}
  \label{tab:uniform}
\end{table}
\subsection{Ablation Studies}
To validate the effectiveness of our proposed framework, we conducted ablation studies on GloTok with a 4096 semantic codebook size and a 12288 visual codebook size.

\subsubsection{The impact of different feature levels}
\begin{figure}[ht]
  \centering
  \includegraphics[width=0.85\linewidth]{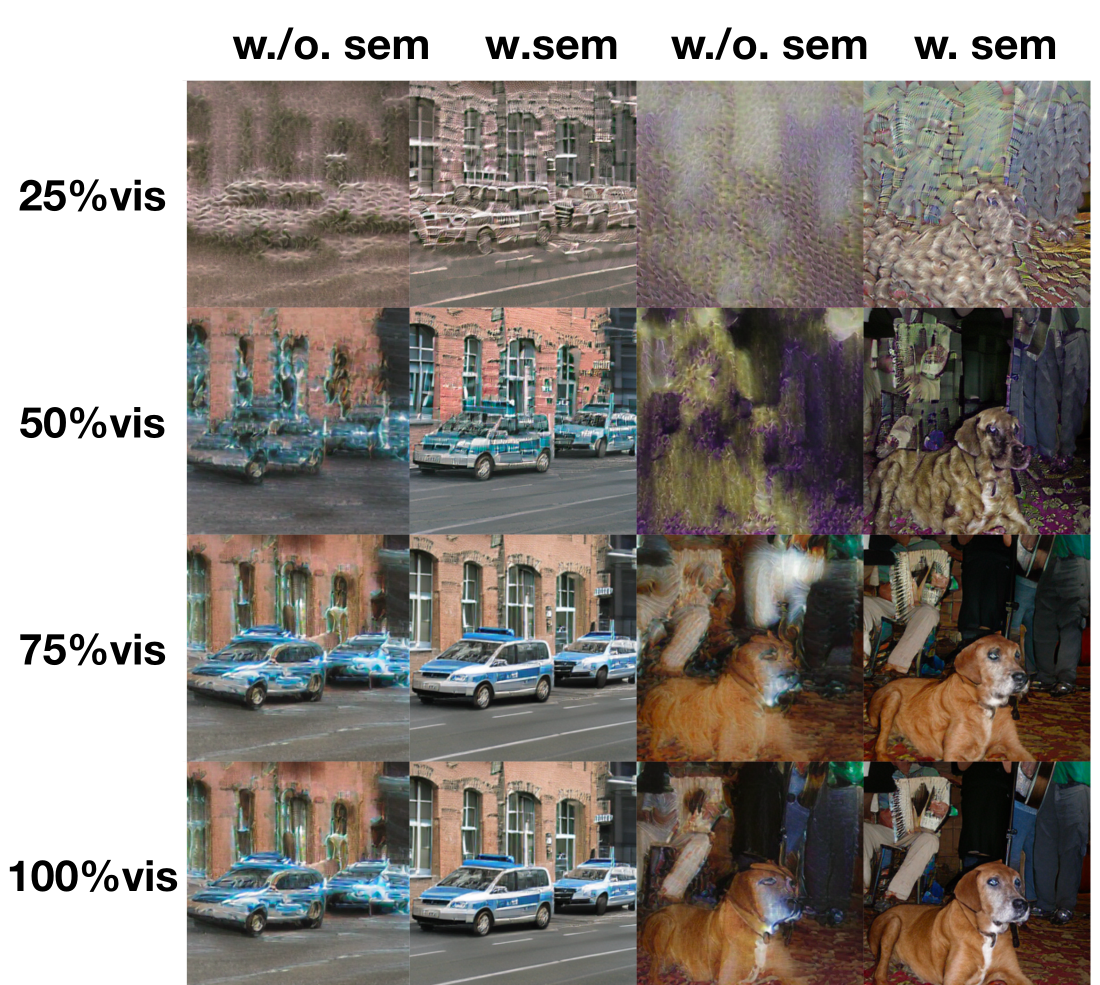}
  \caption{Visualization of the reconstruction on different retention rates of visual features. Each row presents visualization results for visual features with different retention ratios, both with and without semantic features.}
  \label{rec}
\end{figure}

As illustrated in Fig.~\ref{rec}, to evaluate the impact of different feature levels on overall image reconstruction, we visualize how varying retention ratios of visual features (ranging from partial to full preservation) influence image quality both with and without semantic feature guidance.
Specifically, we set the discarded features to 0. 
We retain 100\% of the semantic features while varying the retention ratio of the visual features from 25\% to 100\% to observe their interaction. 

The results show that visual features alone preserve substantial color information in reconstructed images, while structural and textural details are diminished. However, when semantic information is incorporated, the structural and textural components are effectively restored.
The difference indicates that the semantic codebook integrates high-frequency components (e.g., structural and textural elements), while the visual codebook combines low-frequency components (e.g., color distributions). 
Our experiments demonstrate the effectiveness of the global relational learning method in transferring semantics into the codebook.

\subsubsection{The effect of the histogram relation learning}
Our tokenizer is trained for 40 epochs under varying weight configurations of histogram relation learning to investigate its impact on the reconstruction performance of our tokenizer.
As shown in Tab.~\ref{tab:rhist}, our experimental analysis reveals that incorporating the method facilitates a reduction in rFID from 1.43 to 1.21, accompanied by an improvement in PSNR from 22.06 to 22.23.
These findings demonstrate that histogram relation learning plays a pivotal role in enhancing tokenizer reconstruction performance. 
Empirically, setting the hyperparameter weight to 0.01 achieved superior optimization outcomes in our experiments.

\begin{table}[ht]
  \centering
  \begin{tabular}{c|c|ccc}
    \toprule
    \textbf{Histogram Relation} & \textbf{Weight} & \textbf{rFID $\downarrow$} & \textbf{PSNR$\uparrow$} \\
    \midrule
    $\times$ & - & 1.43 & 22.06 \\
    $\checkmark$ & 0.1 & 1.27 & 21.97 \\
    $\checkmark$ & 0.01 & 1.21 & 22.23\\
    \bottomrule
  \end{tabular}
  \caption{Effect of the histogram relation learning. Tokenizers are only trained by the 
 histogram relation learning method without the residual module.}
  \label{tab:rhist}
\end{table}

\subsubsection{Different components}
We evaluated the contributions of individual components to the improvement of GloTok reconstruction.
Results are reported in Tab.~\ref{tab:modules}.
We start with a vanilla dual-codebook design.
Through the adoption of histogram relation learning, our experiments demonstrate a significant improvement in reconstruction quality, with the Fréchet Inception Distance (FID) decreasing from 1.43 to 0.97 and the Peak Signal-to-Noise Ratio (PSNR) increasing from 22.06 to 22.30.
These findings show that histogram relation learning enhances the diversity of the latent space.
We then incorporate the residual learning module, which improves the rFID from 0.97 to 0.92 and improves the PSNR from 22.30 to 22.44. These results demonstrate the effectiveness of the residual learning module in optimizing the reconstruction performance of the tokenizer.

\begin{table}[h]
  \centering
  \begin{tabular}{cc|cc}
    \toprule
    \textbf{Histogram Relation} & \textbf{Residual} & \textbf{rFID $\downarrow$} & \textbf{PSNR $\uparrow$}\\
    \midrule
    $\times$ & $\times$ & 1.43 & 22.06  \\
    $\checkmark$ & $\times$ & 0.97 & 22.30\\
    $\checkmark$ & $\checkmark$ & 0.92 & 22.44  \\
    \bottomrule
  \end{tabular}
  \caption{Ablation study on different components for the improvement of GloTok.}
  \label{tab:modules}
\end{table}

\section{Conclusion}
In this paper, we introduce GloTok, an image tokenizer that leverages global relational information to achieve a more uniform semantic latent distribution. Employing a codebook-wise histogram relation learning method, GloTok effectively regulates the global distribution of tokens across the entire dataset. GloTok does not have to directly access the pre-trained model during training, effectively reducing both training time and GPU memory usage.
In addition, a residual learning module is proposed to mitigate the fine-grained detail lost during discretization.
Experiments on the ImageNet dataset show that GloTok achieves state-of-the-art reconstruction performance and generative performance.

\section*{Acknowledgments}
The computations in this research involved with Yuxi Mi and Shuigeng Zhou were performed using the CFFF platform of Fudan University.

\bibliography{references}

\clearpage 
\appendix
\section*{Supplementary Material}
\section{Data Appendix}
\subsection{Train and validation Dataset}
Our model was trained and validated using ImageNet-1K~\cite{deng2009imagenet}, a commonly adopted dataset for image generation tasks. 

The ImageNet-1K training dataset contains 1,000 classes with a total of 1,288,167 images, while the validation dataset includes 1,000 classes with 50,000 images.

\section{Technical Appendix}
\subsection{Comparison between models using Local Contrastive Loss and GloTok.}
\begin{figure}[hb]
  \centering
  \includegraphics[width=\linewidth]{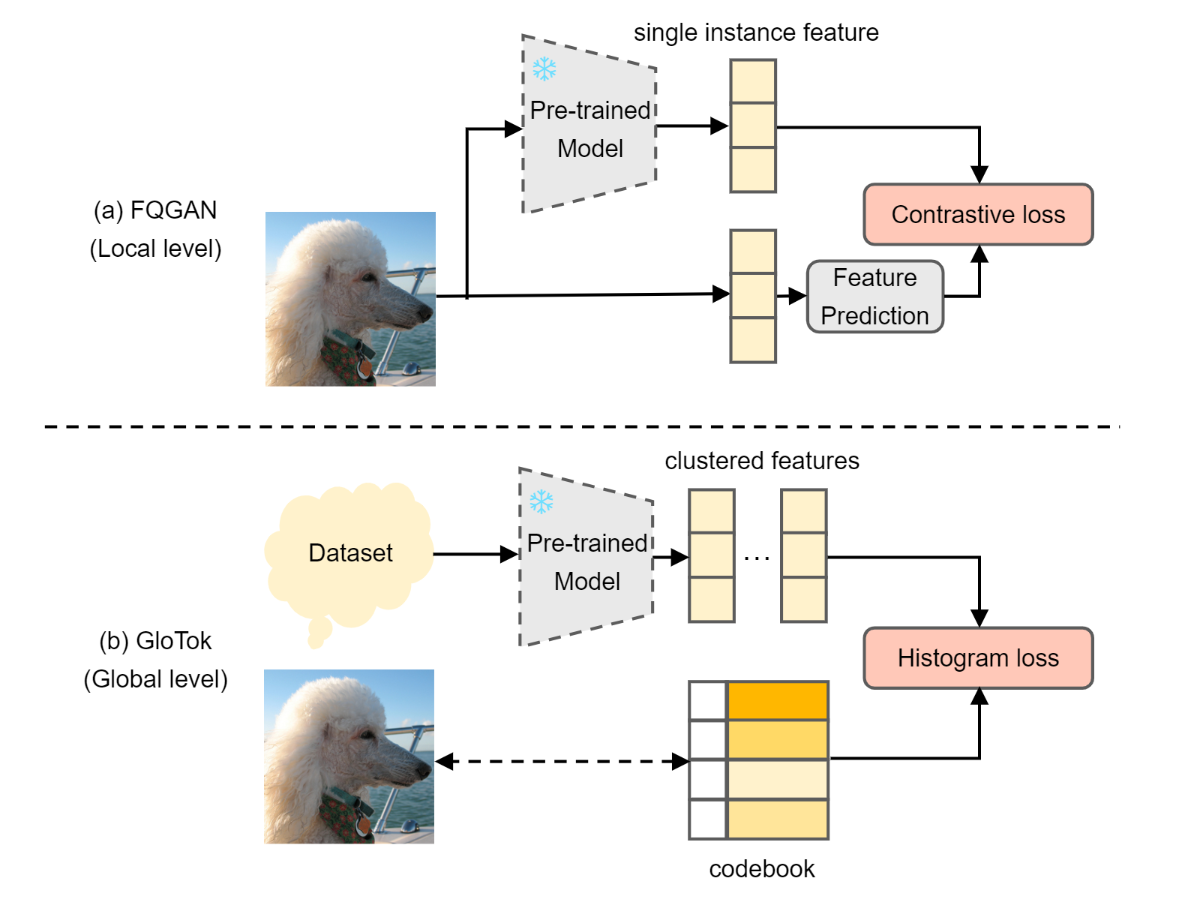} 
  \caption{Framework comparison between SOTAs and GloTok. }
  \label{img:c2s}
\end{figure}
As shown in Tab.~\ref{img:c2s}(a) SOTAs like FQGAN usually adopt a contrastive loss to align discrete features with the features extracted from pre-trained models at the local level. The Tab.~\ref{img:c2s}(b) shows that our GloTok leverages a histogram loss to train the codebook that learns from the clustered features of pre-trained models at the global level.
\subsection{The details of Histogram Relation Learning }
In our work, we adopt a relatively simple histogram relation learning approach to obtain a more uniform semantic feature distribution representation. Alg.~\ref{alg:hist} shows the details of our method. As illustrated in Steps 1–4 of Algorithm 1, we pre-establish prior histogram distributions of pre-trained codebooks or clustered feature before training. In our method, training codebooks using pre-established relational distributions not only enables us to transfer semantic information into the semantic codebook but also accelerates the training process. Our method is computationally simple and efficient, avoiding the need (unlike previous local Contrastive Loss methods) for a pre-trained model to perform forward inference and the training of a separate feature prediction module during the training process. Experimental tests demonstrate that our method reduces training time by over 15\% and decreases GPU memory usage by nearly 10\%.
\begin{algorithm}[h] 
\small
\textbf{Input}:teacher distribution $D_t$, student distribution $D_s$, number of bins $N$, weight factor $\alpha$.\\
\textbf{Output}:histogram relation loss $L_{hist}$.\\
\begin{algorithmic}[1]
\STATE Initialize N bins $B$ from -1 to 1;
\STATE $sim_t$ = $\|$mm($D_t$,$I(B.shape)$) - mm($I(D_t.shape)$,$B$)$\|^2$; \\ 
\COMMENT{I is all-one matrix.}
\STATE $P_t$ = sum(exp($-\alpha \times sim_t$), dim=0, keepdim=True);
\STATE $P_t$ = $P_t$/norm($P_t$, dim=1)
\STATE $sim_s$ = $\|$mm($D_s$,$I(B.shape)$) - mm($I(D_s.shape)$,$B$)$\|^2$;
\STATE $P_s$ = sum(exp($-\alpha \times sim_s$), dim=0, keepdim=True);
\STATE $P_s$ = $P_s$/norm($P_s$, dim=1)
\STATE $L_{hist}$ = KL($\log{P_s}$, $P_t.detach()$);\\ 
\COMMENT{KL is Kullback-Leibler divergence function.}
\end{algorithmic}
\caption{Histogram Relation Learning}
\label{alg:hist}
\end{algorithm}

\subsection{Comparison on different teacher tokens}
We evaluated the impact of different teacher feature space distributions on tokenizer performance, which is shown in Tab.~\ref{tab:teacher}. 
For tokenizers in this experiment, we only employed the histogram relation learning method for training.
We initially conducted experiments on tokens derived from K-means-clustered DINOv2, achieving an rFID of 0.83. 
Moreover, we replicated the same experiment setting on semantic codebooks from dual-codebook tokenizers, observing comparable reconstruction performance.
This experimental result suggests that the source of global relations is not crucial, whereas the effective exploitation of global relational information is what truly matters.

\begin{table}[ht]
  \begin{tabular}{c|c|cc}
    \toprule
    \textbf{Teacher}  & \textbf{Semantics} & \textbf{rFID $\downarrow$} & \textbf{PSNR$\uparrow$} \\
    \midrule
    FQGAN &  CLIP & 0.98 & 22.03\\ 
    ImageFolder & DINOv2 & 0.95 & 22.37 \\
    DINOv2-Kmeans  & DINOv2 & 0.95 & 22.39  \\
    \bottomrule
  \end{tabular}
  \caption{Effect of the source of global relations. The tokenizers in this experiment are not incorporated with residual learning modules.}
  \label{tab:teacher}
\end{table}
\subsection{Comparison residual learning module with RQ-VAE}
We observe that RQ-VAE~\cite{lee2022autoregressive} also adopts a residual-based design, where multiple rounds of quantization are performed on the residuals between quantized features and continuous features to approximate the features lost during quantization. In contrast, our residual learning module employs a transformer block to directly learn the discrepancy between continuous features and current quantized features, thereby avoiding the multiple quantization steps. By decoding the summed features of the learned residual features and quantized features, we find that this process further enhances reconstruction quality. 

As reported in Table 4 of the main text, after incorporating the residual learning module, the reconstruction FID decreases from 0.97 to 0.92, and the PSNR increases from 22.3 to 22.4. This improvement can be interpreted as a further recovery of fine-grained details.

\begin{figure}[t]
  \centering
  \includegraphics[width=0.9\linewidth]{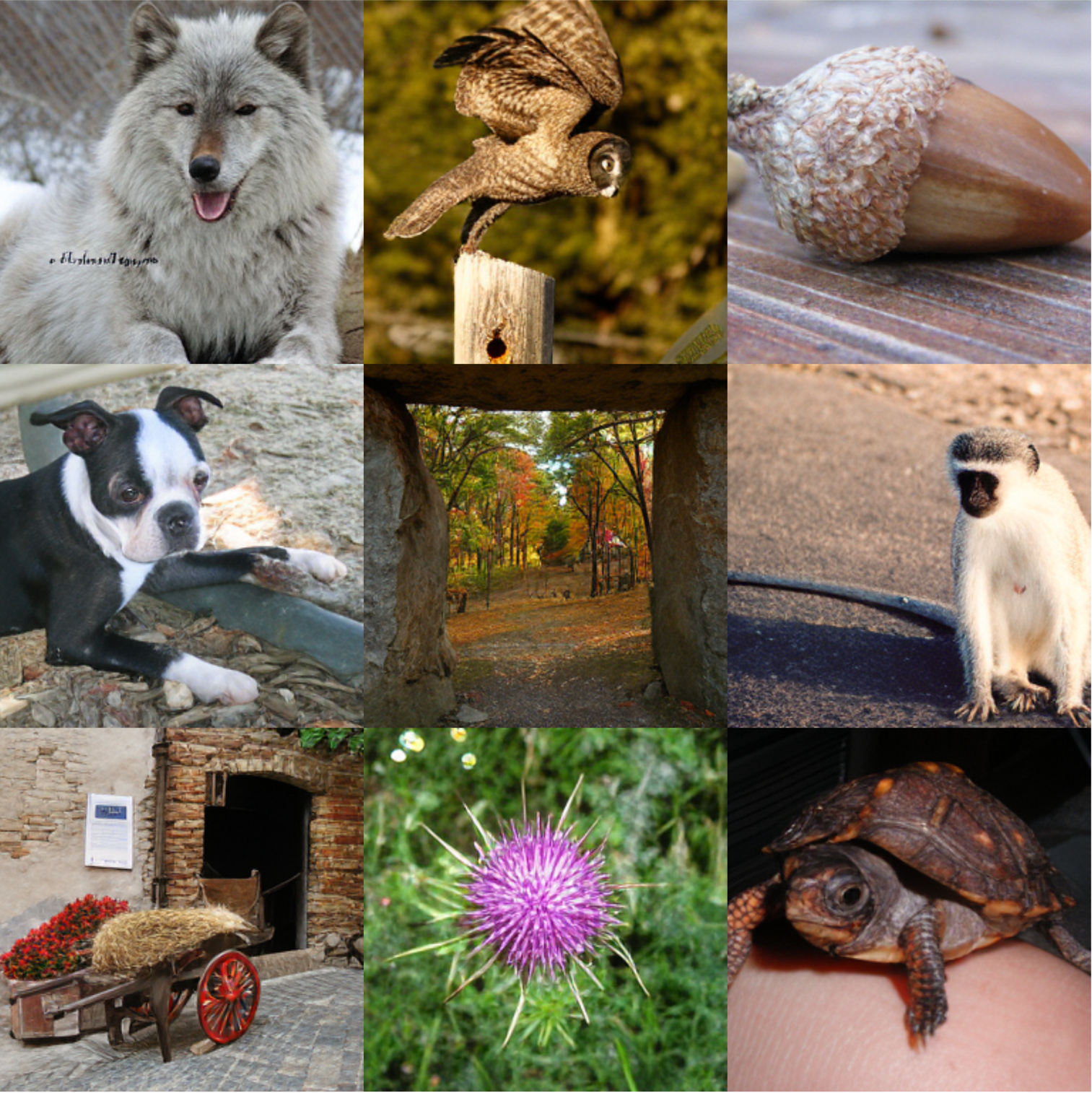} 
  \caption{ImageNet-1k 256$\times$256 reconstruction samples of GloTok on a total codebook size of 16384.}
  \label{fig:rec}
\end{figure}

\subsection{The Reconstruction results of our tokenizer}
For GloTok trained with a visual codebook size of 12288 and a semantic codebook size of 4096, a total 16384 codebook size, our method achieves an FID (Fréchet Inception Distance) of 0.92 and a PSNR (Peak Signal-to-Noise Ratio) of 22.4. Examples of image reconstruction results on our tokenizer are shown in Fig.~\ref{fig:rec}.

\subsection{The Generation results of GloTok trained with FAR-B.}
\begin{figure}[t]
  \centering
  \includegraphics[width=0.9\linewidth]{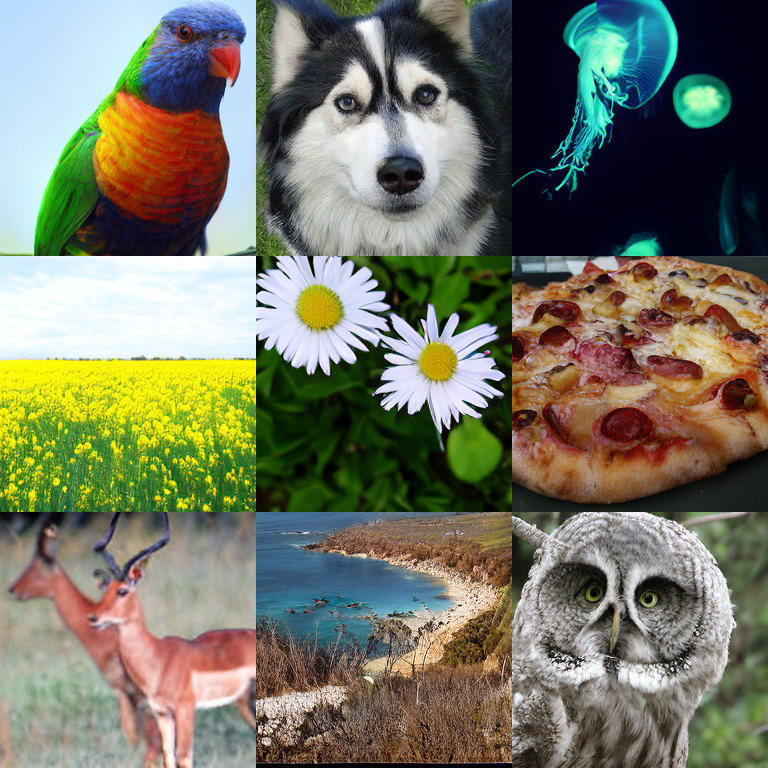} 
  \caption{ImageNet-1k 256$\times$256 generation samples of GloTok with both visual and semantic codebook sizes of 16384.}
  \label{fig:gen-far}
\end{figure}

Our GloTok tokenizer, equipped with both visual and semantic codebook sizes of 16384, achieves a gFID (generalized Fréchet Inception Distance) of 2.95 on the FAR-B benchmark, representing a decrease of 0.43 compared to the architecturally closest FQGAN~\cite{bai2024factorized}. This demonstrates the significant role of our method in enhancing generative performance. Examples of the generated image results produced by FAR-B(415M) are shown in the Fig.~\ref{fig:gen-far}.
\end{document}